%% file: _main.tex
\input{_constants}
\arxiv 

\pdfoutput=1
\documentclass[10pt,twocolumn,letterpaper]{article}
\input{cvpr_header}

\unless\ifarxiv \myexternaldocument{_supplementary} \fi
\definecolor{cvprblue}{rgb}{0.21,0.49,0.74}
\definecolor{mygreen}{RGB}{6, 98, 218}
\begin{document}
\title{Signformer is all you need: Towards Edge AI for Sign Language}

\author{Eta Yang\\
Independent Researcher, Lens AI\thanks{{\href{https://www.lens-ai.net/}{Lens AI}} is an independent research initiative led by the author, currently not a registered legal entity. We release official code and project page available {\href{https://github.com/EtaEnding/Signformer/tree/main}{here}}. } \\
{\tt\small etayangzhourui@gmail.com}}

\maketitle

\input{00_abstract}

\begin{figure}[t]
\centering
\includegraphics[width=\columnwidth]{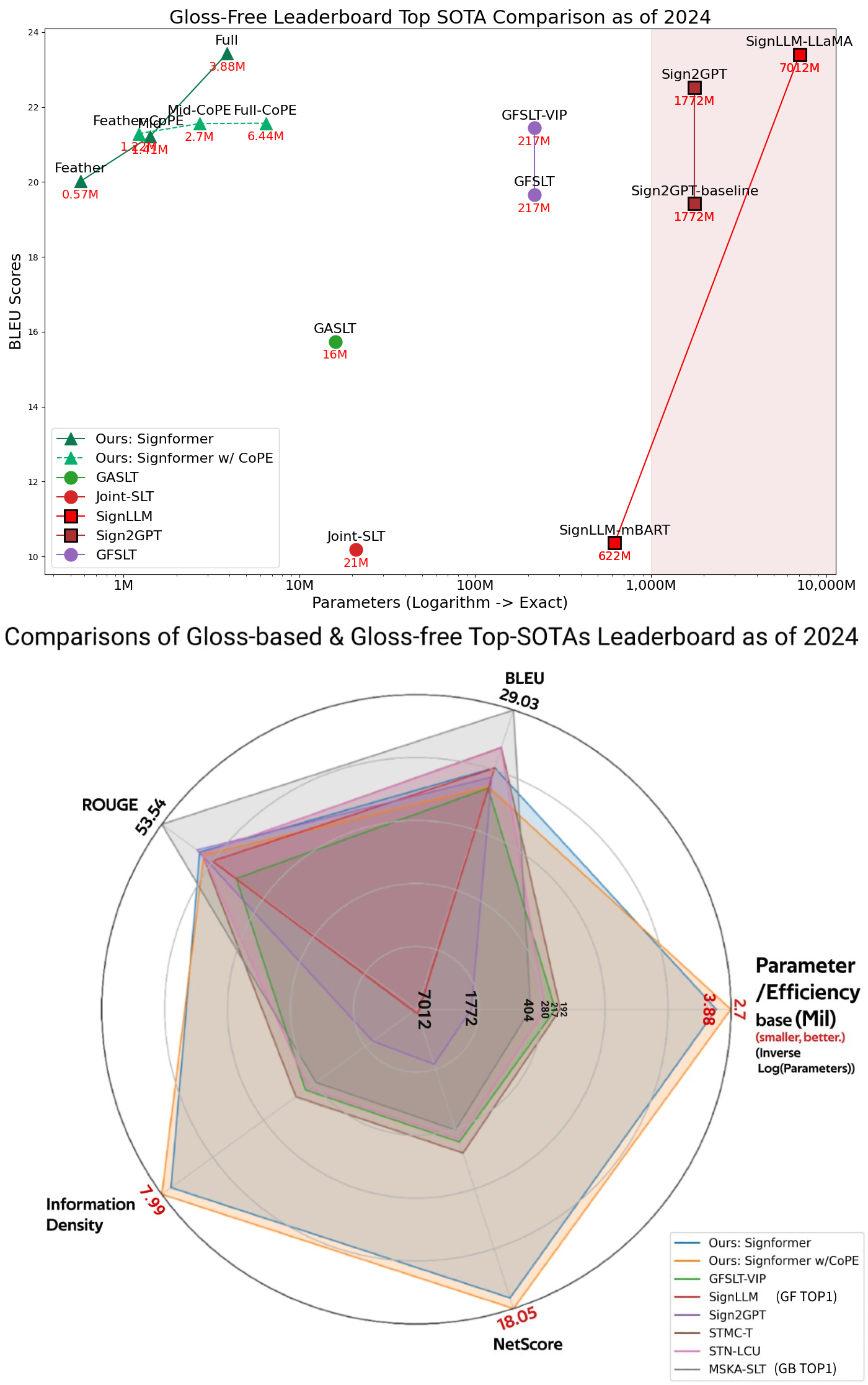}
\caption{\textbf{(TOP): Gloss-Free TOP5 Leaderboard 2024; 
 (Down): Sign Language Translation (Gloss-Based \& Gloss-Free) TOP5 Leaderboard 2024.} Our Signformers, using 3M parameters, achieve the \textbf{\color{red}{2nd}} place on against finests ranged from 1-7B parameters, and
exhibit approximate performance to most SLT gloss-based approaches, while considerably advancing in efficiency, Information Density, and NetScore \cite{netscore}
}
\label{fig:radar}
\end{figure}

\input{01_intro}
\input{02_related}

\input{03_method}

\input{04_experiment}

\input{05_results}
\input{10_conclusion}

{\small
\bibliographystyle{ieeenat_fullname}
\bibliography{11_references}
}

\end{document}


\title{\paperTitle}
\author{\authorBlock}
\maketitlesupplementary

\appendix
\input{12_appendix}

{\small
\bibliographystyle{ieeenat_fullname}
\bibliography{11_references}
}

%% file: _constants.tex
\def\paperID{0000}
\def\confName{CVPR}
\def\confYear{2025}

\newif\ifreview 
\newif\ifarxiv \newcommand{\arxiv}{\arxivtrue}
\newif\ifcamera 
\newif\ifrebuttal 

%% file: cvpr_header.tex
\ifreview \usepackage[review]{cvpr} \fi
\ifarxiv \usepackage[pagenumbers]{cvpr} \fi
\ifrebuttal \usepackage[rebuttal]{cvpr} \fi
\ifcamera \usepackage{cvpr} \fi

\input{_macros}  

\usepackage{xr-hyper}

\makeatletter
\newcommand*{\addFileDependency}[1]{
  \typeout{(#1)}
  \@addtofilelist{#1}
  \IfFileExists{#1}{}{\typeout{No file #1.}}
}

\makeatother
\newcommand*{\myexternaldocument}[1]{
    \externaldocument{#1}
    \addFileDependency{#1.tex}
    \addFileDependency{#1.aux}
}

\definecolor{cvprblue}{rgb}{0.21,0.49,0.74}
\usepackage[pagebackref,breaklinks,colorlinks,allcolors=cvprblue]{hyperref}
\usepackage[capitalize]{cleveref}
\crefname{section}{Sec.}{Secs.}
\crefname{table}{Table}{Tables}
\crefname{figure}{Fig.}{Figs.}

\ifarxiv \crefname{appendix}{App.}{Apps.}
\else \crefname{appendix}{Suppl.}{Suppls.} \fi

%% file: _macros.tex
\usepackage{graphicx}	
\usepackage{amsmath}	
\usepackage{amssymb}	
\usepackage{booktabs}
\usepackage{times}
\usepackage{microtype}
\usepackage{epsfig}
\usepackage{caption}
\usepackage{float}
\usepackage{placeins}
\usepackage{color, colortbl}
\usepackage{stfloats}
\usepackage{enumitem}
\usepackage{tabularx}
\usepackage{xstring}
\usepackage{multirow}
\usepackage{xspace}
\usepackage{url}
\usepackage{subcaption}
\usepackage{xcolor}
\usepackage[hang,flushmargin]{footmisc}

\ifcamera \usepackage[accsupp]{axessibility} \fi

% ...

\ifarxiv  \fi

\newcommand{\R}[1]{{%
    \textbf{%
        \ifstrequal{#1}{1}{\textcolor{red}{R#1}}{%
        \ifstrequal{#1}{2}{\textcolor{blue}{R#1}}{%
        \ifstrequal{#1}{3}{\textcolor{magenta}{R#1}}{%
        \ifstrequal{#1}{4}{\textcolor{teal}{R#1}}{%
                           \textcolor{cyan}{R#1}%
        }}}}%
    }%
}}

%% file: 00_abstract.tex
\begin{abstract}

Sign language translation, especially in gloss-free paradigm, is confronting a dilemma of impracticality and unsustainability due to growing resource-intensive methodologies. Contemporary state-of-the-arts (SOTAs) have significantly hinged on pretrained sophiscated backbones such as Large Language Models (LLMs), embedding sources, or extensive datasets, inducing considerable parametric and computational inefficiency for sustainable use in real-world scenario. Despite their success, following this research direction undermines the overarching mission of this domain to create substantial value to bridge hard-hearing and common populations. Committing to the prevailing trend of LLM and Natural Language Processing (NLP) studies, we pursue a profound essential change in architecture to achieve ground-up improvements without external aid from pretrained models, prior knowledge transfer, or any NLP strategies considered not-from-scratch.

Introducing \textbf{Signformer}, a from-scratch Feather-Giant transforming the area towards Edge AI that redefines extremities of performance and efficiency with LLM-competence and edgy-deployable compactness. In this paper, we present nature analysis of sign languages to inform our algorithmic design and deliver a scalable transformer pipeline with convolution and attention novelty. We achieve new 2nd place on leaderboard with a parametric reduction of 467-1807x against the finests as of 2024 and outcompete almost every other methods in a lighter configuration of 0.57 million parameters.

\end{abstract}

%% file: 01_intro.tex
\section{Introduction}
\label{sec:intro}
Sign language serves as a primary mode of communication for millions of deaf and hard-of-hearing individuals globally. Despite its importance, progress in sign language translation (SLT) has encountered significant obstacles. As a key area within computer vision, SLT is crucial for bridging the communication gap between the deaf community and the hearing population. Traditionally, SLT methods have relied heavily on gloss annotations as intermediate symbolic representations to enhance translation performance. Gloss-based approaches have been effective in simplifying the complex task of translating dynamic visual-gestural languages into written text. However, this methodology is inherently limited by its reliance on gloss annotations, which are resource-demanding and expertise-intensive to produce.

In response to these challenges, the field has shifted focus toward gloss-free sign language translation, aiming to bypass gloss annotations and directly translate sign language videos into text. This new paradigm opens up possibilities for more efficient dataset creation and expansion, but inducing inherent algorithmic complexity for research development, as absence of gloss supervision notably diminishes overall performance \cite{signtransformer}. To our best knowledge as of 2024, most prevailing SOTA methods compensate such drawback by depending on pretrained models, extensive embedding sources like CLIP \cite{clip}, VGG \cite{vgg}, and T5 \cite{t5}, or harnessing sophisticated NLP techniques. Even in recent scenarios, LLMs are perused as backbones to push the boundaries of what is possible. However, these approaches demand unrealistic computational resources, increasing considerable concerns of impracticality and cost. Such direction in research not only diverges from the very fundamental objective of the realm but also hinders its realization towards Edge AI, to facilitate deployable, sustainable SLT solutions that effectively help between the communities.

In light of these challenges, we introduce Signformer, the first groundbreaking Feather-Giant that redefines the gloss-free limit and pivots the realm to Edge AI. Aligning current trend of NLP and LLM research, we position Signformer a from-scratch architectural innovation, establishing a baseline in ultra efficiency and performance. Signformer departs from most methods by forgoing pretrained models, real/pseudo glossary assistance \cite{sign2gpt, signtransformer, SignLLM}, excessive embedding source, extra datasets, or expensive NLP strategies. Leveraging novel convolutions, attention and position encoding mechanisms, we achieve a unprecedented leap in parameter reduction and significant outcompetence against most SOTA competitors,  ranking the \textbf{new 2nd place} across the gloss-free scoreboard as of 2024 as the smallest model across all SLT domain. \textbf{Our flagship Signformer-Full yields better Test Set Performance than the gloss-free Finest SignLLM \cite{SignLLM} while 1807x smaller}, while Signformer-Feather mounted TOP5 with parametric count of 0.57 Million, as demonstrated in \autoref{fig:radar}.

Our contributions are threefold:
\vspace{-0.5pt}
\begin{enumerate}     
\setlength{\itemsep}{0pt}
 \item \textbf{Linguistic Research:} We metricuously analyze the structural nature of diverse sign languages to inform our algorithmic design, generating insightful guidance for future research towards generalizable baselines. \item \textbf{From-Scratch Innovation:} We propose a ultimate-performing-efficient, replicable, end2end, LLM-comparable, and from-scratch architecture that tops gloss-free leaderboard without reliance on external resource-intensive components or complex techniques. \item \textbf{Edge AI Transformation:} We unveil the first Edge-AI-ready baseline in SLT, paving the way for more deployable and scalable applications accessible to the wider population, effectively bridging communication between communities.\end{enumerate}

%% file: 02_related.tex
\section{Related Works}
\subsection{Traditional Approach} 
Early approaches to SLT typically segmented the tasks of recognition and translation into separate stages, with the performance of downstream tasks heavily contingent on the word error rate (WER) \cite{wer} of the preceding stage.  Convolutional neural networks (CNNs) were commonly adopted for visual features, while long-short-term-memory (LSTM) networks handled textual translation. FCN \cite{fcn} was among the first to explore 1D CNNs in Sign Language Recognition (SRT), though its applicaiton was not extend their use to SLT. Compared to transformer-based methods, legacy approaches struggle with capturing long-range dependencies due to the recurrent network's bottleneck. Although methods like CNN-LSTM-HMM \cite{cnnhmm} aim to revise recognition, they fall short in multi-stream effectiveness and underutilize convolutional potential.

\subsection{Transformer-Based Architectures}
The introduction of transformer architectures marked a significant leap forward in SRT as \textbf{Sign Language Transformers (SL-Transformer)} \cite{signtransformer} established a promising baseline of vanilla architecture across the scoreboards by employing a joint-train protocol with gloss supervision that significantly improves translation performance.\par Gloss supervision is then more dedicated to variants on spatial-temporal feature improvement. STMC-Transformer \cite{stmctransformer} harnesses multi-cue analysis, aggregating spatial and temporal channels for better glossary representation and overall translation accuracy. This method follows STMC \cite{stmc} legacy and notably contributes, but demands multiple-model ensemble to realize reported performance. Numerous methods similar to VL-Mapper \cite{vlmapper} distinguish by exploiting additional large-corpus datasets, e.g. Kinetic400 \cite{kinetics400}, WLASL \cite{wlasl}, and CC25 \cite{mbart}, obtained and manually annotated through the Internet and pretrained Language Models like mBART \cite{mbart} or mBERT \cite{mbert}. Nonetheless, exploitation on scaling law and transfer learning hinders genuine architectural reinnovation.

\begin{figure*}[t]
\centering
\includegraphics[width=0.8\textwidth]{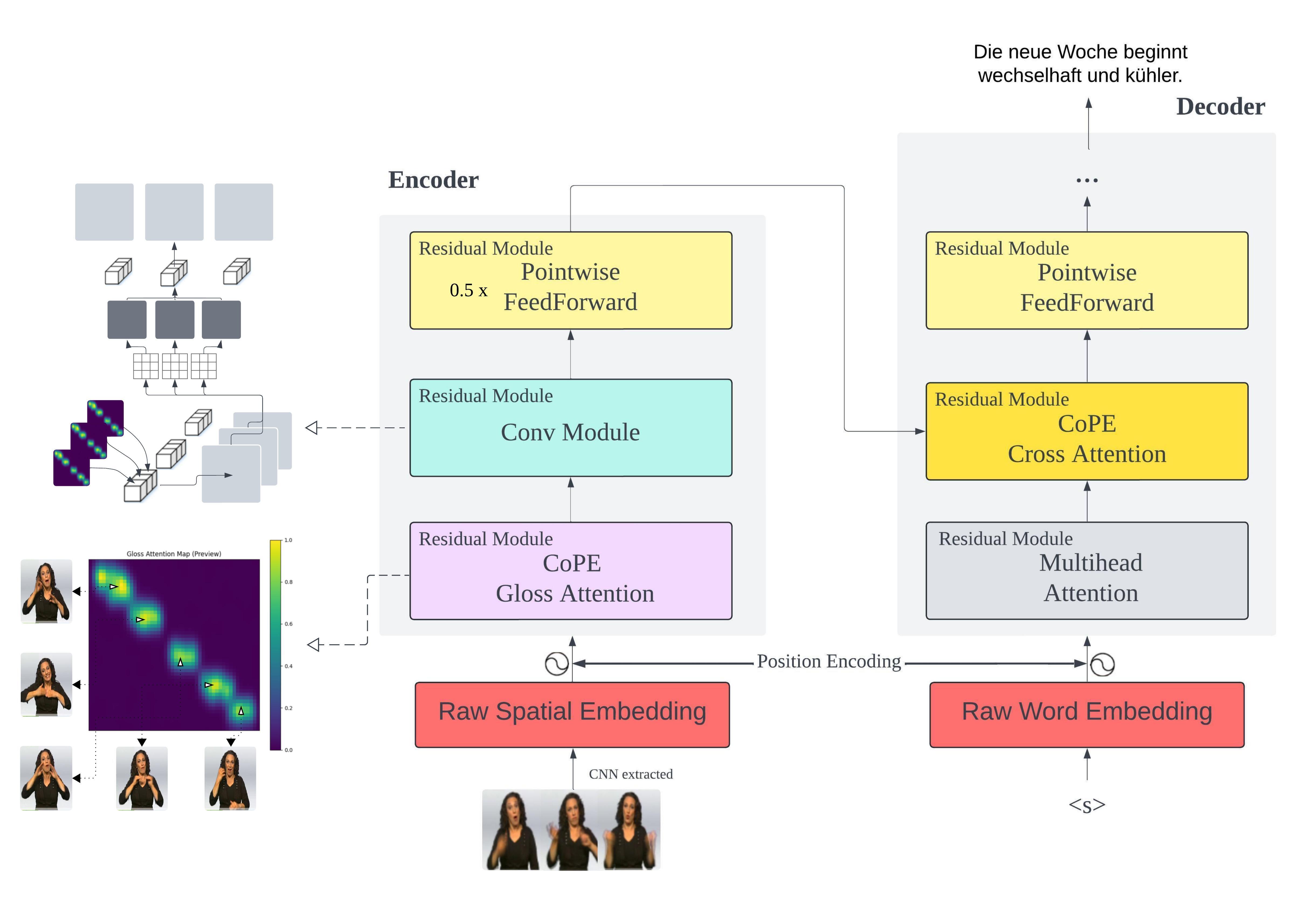}
\caption{\textbf{Architecture of Signformer}, composed of a convolutional module, CoPE-Gloss Attention, and CoPE-Cross Attention. Model is built and trained from-scratch without external embedding or pretrained source, utilizing raw spatial and word embedding layers.}
\label{fig:model overview}
\end{figure*}

\subsection{Gloss-Free Sign Language Translation}
Gloss-Free paradign has appeared the most important shift as ultimate video2text modelling enables more authentic linguistic analysis and deeper insights into natural language use. SL-Transformer (Camgoz et al., 2020) is reproduced by \cite{gaslt} as a vanilla transformer baseline. Successful variants emerge as Zhou et al. (2023) introduce GFSLT-VLP \cite{GFSLT}, a significant advancement in leveraging CLIP to bridge the semantic gap between visual sign language inputs and their corresponding textual outputs. By combining CLIP with masked self-supervised learning, they effectively create a pre-training task that strengthens the model’s ability to learn joint visual-textual representations without contribution of gloss annotations. 

Sign2GPT firstly introduces LLM by integrating a pretrained multilingual GPT, XGLM \cite{xglm} specifically, employed with a LoRA adapter for downstream fine-tuning, after a pretrained encoder on pseudo glosses extracted from datasets. Although successful as the 2nd place in leaderboard, its intensive resource demand raises impractical concerns. By leveraging LLM to another level, SignLLM \cite{SignLLM} has emerged as the 1st performer on the leadeboard of 2024, with a framework combining VQ-Sign and CRA modules. This process of converting video inputs into discrete character-level tokens, aligning them into a word-level structure, and forwarding to an off-the-shell LLama-1 \cite{llama} for translation is computationally severe. Iteratively, visual features are transformed into more “language-like” or pseudo gloss tokens, which pertains to text2text that impact model's generalization and adaptability due to token/prompt abstraction and loss of rich, nuanced, or fine-grained visual subtleties against direct cross-modality modeling. We witness from their report the main BLEU \cite{bleu} contributor is the pretrained LLama 1, while their baseline as reported replaced with a mBART resulted in a substantial performance drop to around 11 from 24.\par

These approaches, while transcending, suffer from severe non-scalability and inefficiency caused by hundreds of millions to billions of parameters. We aim to omit such complex design logic and adhere to direct cross-modal modeling where GASLT's gloss attention \cite{gaslt} encodes gesture information in latent space with greater fine-grained detail preservation since reliance on true/pseudo glosses oversimplifies the rich nature of sign languages, potentially leading to a loss of meaning and context.\par

\begin{figure*}[t]
\centering
\resizebox{\textwidth}{!}{
\includegraphics[]{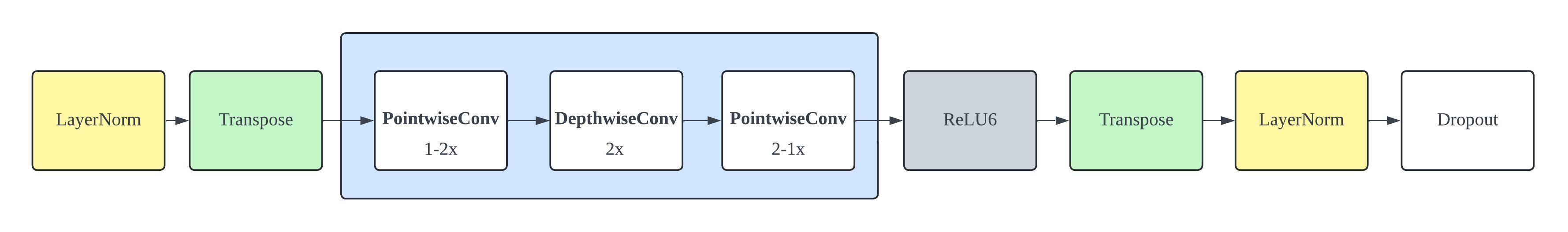}}
\caption{\textbf{Convolution Module,} composed of a stacked Pointwise-Depthwise-Pointwise 1D convolution encapsulated within a symmetric LayerNormalization flow, designed to incorporate with gloss attention to extract higher-level visual features.}
\label{fig:conv overview}
\end{figure*}

%% file: 03_method.tex
\section{Model Description}
Our approach (\autoref{fig:model overview}) follows the baseline SL-Transformer's \cite{signtransformer} training pipeline and its preprocessed dataset, ensuring that our architectural improvements are directly comparable. Features are embedded via a single raw linear layer, while for decoder translation, a raw embedding layer is used.\par

Although it is acknowledged that CLIP, VGG, and other large, pretrained multimodal ViT-based \cite{vit} models offer superior embedding performance, we insist on inheriting original pipeline and starting our model from scratch. This decision aligns with the current trend in research \cite{redundantgpt, efficient1, efficientattention}, which emphasizes demonstrating new language models’ superiority from both innovation and architectural perspectives.

Despite the dominance of models like GPT and LLaMA, convolutional layers have seen underexplored in language models. Prior literature has focused on scaling laws and accumulating multi-modal datasets with tremendous computational resources like H100s or HGX clouds. This unsustainable demand highlights the need for more efficient architectures accessible to the broader research community. 

\subsection{Attention}
We leverage GASLT's \cite{gaslt} novel deformable attention mechanism to fulfill the ultimate end2end cross-modal solution. This method enables us to target localized video segments that exhibit similar temporal features within adjacent frames, attending semantic boundaries of potential glosses within sign language sequences. Departed from textual embedding offered by pseudo-gloss or text2text paradigm, our encoder is capable of comprehending visual-in-nature glosses in latent space while preserving visual nuances essential for accurate understanding of such language towards subsequent modules, under a direct video2text system.\par

Although GASLT incorporates BPEmb \cite{bpemb} to enforce textual embedding and performance by transferring subword information and global sentence-level knowledge, we retain the from-scratch raw embedding layers and training protocols from our mutual ancestor SL-Transformer. We omit this technique to emphasize our model's overall inherent superiority.

\subsection{Convolution Module}
Intuitively, consistent frame progression of signs is naturally considered a 1D-serial tasking whereas 1D convolution outweighs other options. We draw inspiration from Conformer \cite{conformer}, which was predominantly used in automatic speech recognition,  reinvent its convolutional module to obtain more compactness and efficiency tailored for gesture recognition in latent space.\par

By permuting layer normalizations and activations, and adding components to encapsulate the entire convolution process (\autoref{fig:conv overview}), we achieve greater symmetry compared to the original design. This change enhances capturing subtle patterns crucial for accurate gesture understanding. To facilitate high-level feature extraction, we deepen convolution expressiveness with a refined pointwise-depthwise-pointwise 1D convolutional flow to enrich representation of inputs. Unlike the original design, which compresses and halves features between convolution layers using Gated Linear Unit (GLU), our seamless structure preserves significant spatial details identified by gloss attention as legacy utilization hinders crucial throughput. Notably, the depthwise convolution operates across the temporal dimension, effectively modeling the sequential nature, for instance progression and timing of movements, which are critical for distinguishing between different signs within each feature channel.\par
In addition, Swish is removed along with GLU as they could theoretically be combinded into SwishGLU \cite{swishglu}, but according to our empirical evaluations one single RELU6 offers better computational efficiency and even performance. This simplification reduces the computational overhead significantly from old nested activations while yielding higher accuracy for sign translation. Batch normalization is removed as our experiments indicate slight performance worsening. \par
We conducted comparative experiments on the feather configuration against Conformer's original design, and \autoref{tab: Conv ablation} showcased an around 1 boost in BLEU-4 performance across both DEV and TEST sets while maintaining approximate model parameters.

\begin{table}[]
\centering
\resizebox{\columnwidth}{!}{%
\begin{tabular}{ccccc}
\toprule
Conv Module  & Parameters & ROUGE & Dev BLEU-4 & Test BLEU-4 \\ \midrule
Original    & 0.57M       & 46.11& 19.62& 20.24\\
\textbf{Ours} & \textbf{0.57M}       & \textbf{46.24} & \textbf{20.59} & \textbf{20.02}\\
\bottomrule
\end{tabular}%
}
\caption{\textbf{Ablation Studies on Convolution Module}, our reinvention manages to achieve significant BLEU improvements while maintaining same parametric cost}
\vspace*{-5.014mm}
\label{tab: Conv ablation}
\end{table}

\begin{table}[]
\resizebox{\columnwidth}{!}{%
\begin{tabular}{ccccc}
\toprule
Position Encoding& Params & ROUGE & Dev BLEU-4 & Test BLEU-4 \\ \midrule
\textbf{APE}    & \textbf{3.88M} & \textbf{48.53} & \textbf{21.90} & \textbf{23.43}\\
RPE    & 4.00M & 46.41 & 21.27      & 21.33       \\
CoPE   & 6.43M      & 44.82& 19.56& 20.11\\
\bottomrule
\end{tabular}%
}
\caption{\textbf{Ablation Studies on optimal Position Encoding (Single)}, experimented with Multihead Attention module since Gloss Attention doesn't support RPE, and APE fulfills our theories and minimizes parametric cost}
\vspace*{-10pt}
  \label{tab: PE ablation}
\end{table}

\subsection{Position Encoding}
Unfortunately, the majority of current research and SOTA papers on sign language fail to thoroughly analyze its systemic architecture. Much of the existing works focus primarily on engineering experiments and the publication of deep learning models, often overlooking the linguistic foundations of sign language. As a critical branch of linguistics, sign language requires careful consideration of appropriate algorithms, which many studies neglect. In this section, we will provide an in-depth analysis of the linguistic systems associated with publicly available datasets, thereby enabling a more informed investigation into the applicability of position encoding.\par

In the realm of linguistics, the importance of word order varies influentially among languages, particularly when comparing German with mainstream languages including English, Japanese, and Chinese. German, which primarily follows a Subject-Object-Verb (SOV) structure in subordinate clauses and a Subject-Verb-Object (SVO) structure in main clauses \cite{sov2, sov3, svo1}, places considerable emphasis on the order of words to convey meaning accurately. This duality in structure necessitates strict adherence to word order to avoid ambiguity, especially since the positioning of verbs and objects can alter the intended message. Similarly, English and Chinese adhere strictly to an SVO order, where the subject precedes the verb, and the object follows. This rigid structure makes the order of words crucial for understanding, as deviations can lead to misunderstandings or a complete change in meaning. For instance, \textit{\textbf{The cat chased the dog}} is distinctly different from \textit{\textbf{The dog chased the cat}}, highlighting the reliance on word order to convey syntactic relationships clearly. In systems with flexible Syntax, like Japanese (SOV but flexible) \cite{JSLSign} and American Sign Language (Variable Word Order)\cite{ASlsign}, non-manual signals, e.g special expressions, head movements, etc., and context become more crucial in interpretation.\par

Upon replacement of attention, the original relative position encoding (RPE) \cite{transformerxl} embedded within multi-head attention is removed. As we follow frameworks of mainstream languages and evaluate upon the German dataset Phoenix14T \cite{phoenix14t}, RPE introduces challenges in translation. Besides extensive computational overhead and difficulty in maintaining strict word order, RPE's algorithm may potentially lead to ambiguity since it is indistinguishable between sentences with similar words but different orders.\par

By assigning a fixed positional value to each token, Absolute Position Encoding (APE) \cite{transformer} ensures that each word's position is explicitly and uniquely represented, aligning better with mainstream grammar structures along with gloss attention. We conducted experiments to compare the uses of different encodings, and the results, as detailed in \autoref{tab: PE ablation}, justify our theory-based projections, indicating that APE conforms to mainstream languages better since highly inflected or context-reliant factors or features can be less critical whereas strict word order is the primary means of conveying syntactic relationships. 

\subsection{Decoder Attention}
Regardless of various workarounds to enhance embedding expressiveness to support textual generation, we opted to use the original transformer decoder to justify our architectural efficacy and potential for future upgrades. The rise of LLMs has provided various techniques to reinforce text generation, yet most SOTA studies rely on pretrained LLMs, extra datasets, data augmentation, and workarounds for extensive competence, inducing considerable cost and complexity to hinder focusing on essential architectural optimizations. Though unfair competition for us, using a raw transformer decoding algorithm ensures an extendable baseline for future enhancements as trick integration is linearly effective. \par

\begin{table}[]
\centering
\resizebox{\columnwidth}{!}{%
\begin{tabular}{cccccc}
\toprule
Decoder                          & Encoder         & Parameters & ROUGE & Dev BLEU-4 & Test BLEU-4 \\ \midrule
\multirow{2}{*}{Cross Attention} & Gloss Attention & 0.57M      & 46.24& 20.59& 20.02\\
& + w/ CoPE       & 0.9M       & 45.77 & 19.40      & 19.83       \\ \midrule
\multirow{2}{*}{+ w/ CoPE}       & Gloss Attention & 0.9M       & 47.02 & 20.46      & 20.13       \\
 & + w/ CoPE       & \textbf{1.22M} & \textbf{47.25}& \textbf{20.01}&  \textbf{21.29}\\ \bottomrule
\end{tabular}%
}
\caption{\textbf{Integration of CoPE with attentions}, ablation results (64 HiddeneSize) indicate that the dual applications deliver the highest BLEU score improvements, with Cross Attention being the primary contributor to this enhancement.}
\vspace*{-10pt}
  \label{tab: CoPE ablation}
\end{table}

\subsubsection{Contextual Position Encoding}
While APE established a global scale, stable positional framework imperative for modeling overall sequential structure, CoPE \cite{cope} instead makes a vital contribution by offering fine-grained, context-sensitive adjustments to positional encoding at a local scale that APE alone cannot provide. CoPE effectively manages nuanced features, i.e. non-manual makers such as facial expressions and head movements. It also excels in handling particles (Japanese), inflections (Latin), and syntactic structures like Topic-Comment patterns in sign languages, where glossary ordering can be flexible. In this case, for more rigid languages like German, Chinese, and English, CoPE primarily captures these nuanced signals. 
\par
Also, while Cross Attention typically attends the entire input sequence, we are adding CoPE as a layer of locality to how attention is distributed, identifying specific regions within the sequence as particularly relevant to the current output token. This localized focus within the broader global context enables the model to prioritize the most pertinent parts of the input sequence at each decoding step.
\par

As our convolutional module specializes in extracting more locality and finer features, from the input sequence, a rich, hierarchical representation is encoded into a latent space. This synergy between the nuanced feature extraction provided by convolutional layers and the cross attention of context-aware positional adjustments facilitated by CoPE allows the decoder to maintain both the broad structural integrity (supported by APE) and capture both direct relationships between adjacent signs and more complex, context-dependent patterns. Our combination of 2 position encodings leads to higher BLEU scores, particularly witnessed in lighter-weight models where the careful handling of both local and global dependencies is essential. In addition, without explicit gloss annotations, CoPE helps mitigate the positional ambiguity that can arise from APE against sequences of repetitive patterns. By allowing the encoder to condition positional encoding on the surrounding latent glosses, CoPEs ensure that the gloss attention mechanism can more precisely align input signs with their appropriate positions in the output sequence, leading to more accurate and fluent translations, as shown in the qualitative section.\par

However, we witness negativity of CoPE in application with sole attention for which reason we hypothesize that only dual applications create a feedback loop where encoder attention augments positional focus, and decoder attention peruses these dynamical refinements for a more comprehensive and synergistic reinforcement of positional adjustments across the entire model during alignment.\par

\begin{table}[]
\centering
\resizebox{\columnwidth}{!}{%
\begin{tabular}{cccccc}
\toprule
\multicolumn{1}{c}{Hidden Size}          & \multicolumn{1}{c}{CoPE} & \multicolumn{1}{c}{Params} & \multicolumn{1}{c}{ROUGE}          & \multicolumn{1}{c}{Dev BLEU-4}     & Test BLEU-4    \\ \midrule
\multicolumn{1}{c}{\multirow{2}{*}{32}}  & \multicolumn{1}{c}{yes}  & \multicolumn{1}{c}{0.58M}  & \multicolumn{1}{c}{\textbf{44.11}} & \multicolumn{1}{c}{\textbf{17.34}} & \textbf{17.87} \\
\multicolumn{1}{c}{}                     & \multicolumn{1}{c}{no}   & \multicolumn{1}{c}{0.26M}  & \multicolumn{1}{c}{43.54}          & \multicolumn{1}{c}{16.92}          & 16.82          \\ \midrule
\multicolumn{1}{c}{\multirow{2}{*}{\textbf{64}}}  & \multicolumn{1}{c}{yes}  & \multicolumn{1}{c}{1.22M}  & \multicolumn{1}{c}{\textbf{47.25}} & \multicolumn{1}{c}{\textbf{20.01
}} & \textbf{21.29}\\
\multicolumn{1}{c}{}                     & \multicolumn{1}{c}{no}   & \multicolumn{1}{c}{0.57M}  & \multicolumn{1}{c}{46.24}          & \multicolumn{1}{c}{20.59}          & 20.02\\ \midrule
\multicolumn{1}{c}{\multirow{2}{*}{96}}  & \multicolumn{1}{c}{yes}  & \multicolumn{1}{c}{1.91M}  & \multicolumn{1}{c}{\textbf{48.04}} & \multicolumn{1}{c}{\textbf{21.38}} & \textbf{20.92} \\
\multicolumn{1}{c}{}                     & \multicolumn{1}{c}{no}   & \multicolumn{1}{c}{0.96M}  & \multicolumn{1}{c}{47.47}          & \multicolumn{1}{c}{21.20}          & 20.51          \\ \midrule
\multicolumn{1}{c}{\multirow{2}{*}{\textbf{128}}} & \multicolumn{1}{c}{yes}  & \multicolumn{1}{c}{2.70M}   & \multicolumn{1}{c}{\textbf{47.98}} & \multicolumn{1}{c}{\textbf{21.82}} & \textbf{21.56} \\
\multicolumn{1}{c}{}                     & \multicolumn{1}{c}{no}   & \multicolumn{1}{c}{1.41M}  & \multicolumn{1}{c}{47.15}          & \multicolumn{1}{c}{21.49}          & 21.21          \\ \midrule
\multicolumn{6}{c}{......}                                                                                                                                                                     \\  \midrule
\multicolumn{1}{c}{\multirow{2}{*}{\textbf{256}}} & \multicolumn{1}{c}{yes}  & \multicolumn{1}{c}{6.44M}  & \multicolumn{1}{c}{47.41} & \multicolumn{1}{c}{21.60} & 21.57 \\ 
\multicolumn{1}{c}{}                     & \multicolumn{1}{c}{no}   & \multicolumn{1}{c}{3.88M}   & \multicolumn{1}{c}{\textbf{48.53}} & \multicolumn{1}{c}{\textbf{21.92}} & \textbf{23.43}\\ \bottomrule
\end{tabular}%
}
\caption{\textbf{Ablation Studies on CoPE and HiddenSize scalability}, we witness significant BLEU improvement on lightweight configrations, but negativeness if HiddenSize above 128. We neglect configurations worsening performance as that's trivial to present.}
\vspace*{-10pt}
\label{tab:ablation_cope_hiddensize}
\end{table}

%% file: 04_experiment.tex
\section{Experiment}
We evaluate Signformers using SLT's primary benchmark PHOENIX14T German dataset \cite{phoenix14t}. And due to the access restrictions of CSL-Daily adminstration, which is only available to academic institutions or research laboratories, we were unable to include this dataset in our evaluation. However, as explained earlier with our deep research in sign language structures, we have ensured both datasets are representative and consistent in an
ture and generalizability of our model across different languages and contexts, providing a robust evaluation within the scope of available resources.

\begin{table*}[]
\centering
\resizebox{\linewidth}{!}{%
\begin{tabular}{cccccc}
\toprule
\multicolumn{1}{c}{Method} &
  \multicolumn{1}{c}{Extra Technique} &
  \multicolumn{1}{c}{Params (M)} &
  \multicolumn{1}{c}{ROUGE} &
  \multicolumn{1}{c}{Dev BLEU-4} &
  Test BLEU-4 \\ \midrule
\multicolumn{6}{c}{\cellcolor[rgb]{0.753,0.753,0.753}  \textbf{Gloss-Based}} \\ \midrule
\multicolumn{1}{c}{SL-Transformer \cite{signtransformer}} &
  \multicolumn{1}{c}{Scratch} &
  \multicolumn{1}{c}{21} &
  \multicolumn{1}{c}{47.26} &
  \multicolumn{1}{c}{22.38} &
  21.32 \\
\midrule
\multicolumn{1}{c}{} &
  \multicolumn{1}{c}{Scratch} &
  \multicolumn{1}{c}{} &
  \multicolumn{1}{c}{46.98} &
  \multicolumn{1}{c}{21.78} &
  21.68 \\

\multicolumn{1}{c}{\multirow{-2}{*}{BN-TIN-Tran.+SignBT \cite{bintintrans}}} &
  \multicolumn{1}{c}{Dataset; Pretrain; DataAug} &
  \multicolumn{1}{c}{\multirow{-2}{*}{-}} &
  \multicolumn{1}{c}{49.54} &
  \multicolumn{1}{c}{24.45} &
  24.32 \\
  \midrule
\multicolumn{1}{c}{} &
  \multicolumn{1}{c}{Pretrained; Transfer Learning} &
  \multicolumn{1}{c}{65*} &
  \multicolumn{1}{c}{46.77} &
  \multicolumn{1}{c}{22.47} &
  24.00 \\
\multicolumn{1}{c}{\multirow{-2}{*}{STMC-T \cite{stmctransformer}}} &
  \multicolumn{1}{c}{+ Ensembling, 8 Models} &
  \multicolumn{1}{c}{192*} &
  \multicolumn{1}{c}{48.24} &
  \multicolumn{1}{c}{24.68} &
  25.40 \\
  \midrule
\multicolumn{1}{c}{} &
  \multicolumn{1}{c}{Pretrained Embedding} &
  \multicolumn{1}{c}{35} &
  \multicolumn{1}{c}{-} &
  \multicolumn{1}{c}{23.23} &
  23.65 \\
\multicolumn{1}{c}{\multirow{-2}{*}{STN-LCU \cite{stnlcu}}} &
  \multicolumn{1}{c}{+ Ensembling, 8 Models} &
  \multicolumn{1}{c}{280} &
  \multicolumn{1}{c}{-} &
  \multicolumn{1}{c}{25.59} &
  25.33 \\
  \midrule
\multicolumn{1}{c}{} &
  \multicolumn{1}{c}{Scratch} &
  \multicolumn{1}{c}{} &
  \multicolumn{1}{c}{46.67} &
  \multicolumn{1}{c}{20.70} &
  21.36 \\
\multicolumn{1}{c}{\multirow{-2}{*}{VL-Mapper \cite{vlmapper}}} &
  \multicolumn{1}{c}{Extra 3 Datasets; Pretrained} &
  \multicolumn{1}{c}{\multirow{-2}{*}{610}} &
  \multicolumn{1}{c}{53.10} &
  \multicolumn{1}{c}{27.61} &
  28.39 \\
  \midrule
 MSKA-SLT \cite{mskaslt} & 3D, Pretrained; LM; Embedding& 404& 54.04& 27.63&29.03\\ \midrule
\multicolumn{6}{c}{\cellcolor[HTML]{C0C0C0} \textbf{Gloss-Free}} \\ \midrule
\multicolumn{1}{c}{Conv2d-RNN \cite{cnnhmm}} &
  \multicolumn{1}{c}{Scratch} &
  \multicolumn{1}{c}{-} &
  \multicolumn{1}{c}{31.00} &
  \multicolumn{1}{c}{9.12} &
  8.35 \\
  \midrule
\multicolumn{1}{c}{SL-Transformer \cite{gaslt}} &
  \multicolumn{1}{c}{Scratch} &
  \multicolumn{1}{c}{21} &
  \multicolumn{1}{c}{31.10} &
  \multicolumn{1}{c}{-} &
  10.19 \\
  \midrule
\multicolumn{1}{c}{Tokenization SLT \cite{tokenizationslt}} &
  \multicolumn{1}{c}{Multi-Task; Knowledge Transfer} &
  \multicolumn{1}{c}{-} &
  \multicolumn{1}{c}{36.28} &
  \multicolumn{1}{c}{-} &
  13.25 \\
  \midrule
\multicolumn{1}{c}{TSPNet \cite{tspnet}} &
  \multicolumn{1}{c}{Scratch} &
  \multicolumn{1}{c}{-} &
  \multicolumn{1}{c}{34.96} &
  \multicolumn{1}{c}{-} &
  13.41 \\
  \midrule
\multicolumn{1}{c}{GASLT \cite{gaslt}} &
  \multicolumn{1}{c}{Knowledge Transfer; Embedding} &
  \multicolumn{1}{c}{16} &
  \multicolumn{1}{c}{39.86} &
  \multicolumn{1}{c}{-} &
  15.74 \\ \midrule
\multicolumn{1}{c}{} &
  \multicolumn{1}{c}{Pretrained; LM; Data Aug; Knowledge} &
  \multicolumn{1}{c}{} &
  \multicolumn{1}{c}{40.93} &
  \multicolumn{1}{c}{19.84} &
  19.66 \\
\multicolumn{1}{c}{\multirow{-2}{*}{GFSLT-VIP \cite{GFSLT} }} &
  \multicolumn{1}{c}{+ Embedding; Transfer Learning} &
  \multicolumn{1}{c}{\multirow{-2}{*}{217}} &
  \multicolumn{1}{c}{43.72} &
  \multicolumn{1}{c}{22.12} &
  21.44 \\
  \midrule
\multicolumn{1}{c}{} &
  \multicolumn{1}{c}{LM} &
  \multicolumn{1}{c}{} &
  \multicolumn{1}{c}{-} &
  \multicolumn{1}{c}{11.95} &
  10.36 \\
\multicolumn{1}{c}{\multirow{-2}{*}{\textbf{SignLLM} \cite{SignLLM}}} &
  \multicolumn{1}{c}{+ \textbf{Pretraining; LLM}} &
  \multicolumn{1}{c}{\multirow{-2}{*}{\textbf{7012}}} &
  \multicolumn{1}{c}{\textbf{46.88}} &
  \multicolumn{1}{c}{\textbf{25.25}} &
  \textbf{23.40} \\
  \midrule
\multicolumn{1}{c}{} &
  \multicolumn{1}{c}{Scratch but LLM} &
  \multicolumn{1}{c}{} &
  \multicolumn{1}{c}{45.23} &
  \multicolumn{1}{c}{-} &
  19.42 \\
\multicolumn{1}{c}{\multirow{-2}{*}{Sign2GPT \cite{sign2gpt}}} &
  \multicolumn{1}{c}{Pseudo-Gloss Pretraining; LLM} &
  \multicolumn{1}{c}{\multirow{-2}{*}{1772}} &
  \multicolumn{1}{c}{48.90} &
  \multicolumn{1}{c}{-} &
  22.52 \\ \midrule
\multicolumn{6}{c}{\cellcolor[HTML]{C0C0C0} \textbf{Ours (Gloss-Free)}} \\ \midrule
\multicolumn{1}{c}{Signformer - Feather} &
  \multicolumn{1}{c}{} &
  \multicolumn{1}{c}{0.57} &
  \multicolumn{1}{c}{46.24} &
  \multicolumn{1}{c}{20.59} &
  20.02 \\
\multicolumn{1}{c}{\textbf{{\color[HTML]{FE0000}Signformer - F w/ CoPE}}} &
  \multicolumn{1}{c}{} &
  \multicolumn{1}{c}{{\color[HTML]{FE0000} \textbf{1.22}}} &
  \multicolumn{1}{c}{{\color[HTML]{FE0000} \textbf{47.25}}} &
  \multicolumn{1}{c}{{\color[HTML]{FE0000} \textbf{20.01}}} &
  {\color[HTML]{FE0000} \textbf{21.29}}\\
\multicolumn{1}{c}{Signformer - Mid} &
  \multicolumn{1}{c}{} &
  \multicolumn{1}{c}{1.41} &
  \multicolumn{1}{c}{47.15} &
  \multicolumn{1}{c}{21.49} &
  21.21 \\
\multicolumn{1}{c}{\textbf{{\color[HTML]{FE0000}Signformer - M w/ CoPE}}} &
  \multicolumn{1}{c}{} &
  \multicolumn{1}{c}{{\color[HTML]{FE0000} \textbf{2.70}}} &
  \multicolumn{1}{c}{{\color[HTML]{FE0000} \textbf{47.98}}} &
  \multicolumn{1}{c}{{\color[HTML]{FE0000} \textbf{21.82}}} &
  {\color[HTML]{FE0000} \textbf{21.56}} \\
\multicolumn{1}{c}{\textbf{{\color[HTML]{FE0000}Signformer - Full}}} &
  \multicolumn{1}{c}{} &
  \multicolumn{1}{c}{{\color[HTML]{FE0000} \textbf{3.88}}} &
  \multicolumn{1}{c}{{\color[HTML]{FE0000} \textbf{48.53}}} &
  \multicolumn{1}{c}{{\color[HTML]{FE0000} \textbf{21.92}}} &
  {\color[HTML]{FE0000} \textbf{23.43}} \\
\multicolumn{1}{c}{Signformer - F w/ CoPE} &
  \multicolumn{1}{c}{\multirow{-6}{*}{Scratch}} &
  \multicolumn{1}{c}{6.44} &
  \multicolumn{1}{c}{47.41} &
  \multicolumn{1}{c}{21.60} &
  21.57 \\
  \bottomrule
\end{tabular}%
}
\caption{
\textbf{Quantitave Comparions on Gloss-based \& Gloss-free state-of-the-art.}
All parameter info is acquired directly from authors through emails, otherwise from official repositories or papers. \\
We thoroughly document extra techniques considered non-scratch, as well as the reported baselines for each SOTA model. \textbf{+} indicates adding upon the described in the row above. \\
\textbf{*} denotes an estimate based on the paper’s report. We confirm that STMC has 46.4M parameters as communicated by the original authors. SMTC-T employs a 2-layer transformer with a hidden size of 512 and 2048 feedforward units, using 8 models for ensembling.}
\label{tab: sota}
\end{table*}

\subsection{Implementation and Training Details}
In our gloss-free sign language translation study, we meticulously configured our model with a hidden size of 256 or below for all layers — any above found detrimental — and a kernel size of 31 for the convolution modules. A batch size of 32 was primarily utilized during training. We employed the SophiaG \cite{sophiag} optimizer, integrating gradient clipping and adaptive learning rates to enhance training stability and performance. The initial learning rate was set to 0.004 and was dynamically adjusted using a plateau scheduler, which we identified as optimal for maximizing model capability.
Throughout our experiments, we observed that models with very small hidden size became quite sensitive to SophiaG, resulting in unstable performance variability across different runs. Consequently, we experimented with different optimizers to acquire the best configuration results. Specifically, for Mid and Full configurations, including CoPE, with hidden sizes above 128, we utilized SophiaG. For the Feather models, which has a hidden size of only 64, we adopted AdamW \cite{adamw} to maintain stability. Simultaneously, layer deepening generally offers diminishing returns in terms of the performance-to-parameter ratio against hidden size enlargement. Accordingly, in our comparisons with other works, we chose to feature configurations offering tangible benefits only. Our model typically converged within 25-35 epochs, achieving its highest accuracy performance within this range. 

\subsection{Ablation Studies}

Originally, we introduced CoPE as prime position encoding to replace APE and RPE to examine modelling based on German syntatic nature, and it has been observed to perform worst on the German dataset. This underperformance substantiates the research and our analysis that APE defends the fundamental solution to framework of rigidity but CoPE as addition develops further layers of locality. 

Ablation testing, as detailed in \autoref{tab: CoPE ablation}, revealed that CoPE’s main advantage arises from cross-modal alignment, as solitary integration with gloss attention impairs performance while dual application significantly boosts BLEU and ROUGE \cite{rouge} scores by approximately 1.27 and 1.01, respectively, on the Test Set. This algorithmic fusion leads to a marked improvement, addressing the bottleneck caused by reduced model size. However, performance declines as the hidden size increases beyond 128 (\autoref{tab:ablation_cope_hiddensize}), which we speculate is due to CoPE’s redundant complexity against the language’s nuanced features which can be resolved with a smaller scale.\par

Despite the current limitations posed by limited modality of RGB image and the linguistic characteristics of existing dataset, the future holds potential for CoPE to excel on a larger scale. As global languages expand and richly annotated datasets encompassing heterogeneous data types become more accessible, CoPE’s context-dependent adaptability could provide significant advantages, especially for flexible ones such as Japanese, Korean, French, and others.

%% file: 05_results.tex
\section{Results}
\subsection{Quantitative Comparison}
We established communication with authors to reach out in-depth information excluded from papers, specifically parameters. For comprehensive comparisons across all scoreboards, we utilize metrics such as BLEU-4, ROUGE, and parameters to compare our Signformer lineups and other leading competitors on the PHOENIX14T dataset.
As shown in \autoref{tab: sota}, the translation performance of proposed Signformer has succeeded in substantial benchmark superiority and pushed extremity of compactness footprint that's pivotal across all competitors. We present statistics for various configuration with/out CoPE. Our Signformer-Full accomplished a remarkable leap from 10.31 to 23.43 (227.3\%) in BLEU-4 and streamlined capacity from 7012 million (7B) to 3.88 million parameters (1807x smaller). Signformer-Full is comparable against gloss-free finest SignLLM; meanwhile Signformer-Feather reached beyond 20 in BLEU-4, outperforming the baseline of GFSLT-VIP with only 0.57 million parameters. Its upgraded version with CoPE of 1.2 million delivers approximate results against complete GFSLT-VIP.\par
To preview model capability more comprehensively against gloss-based models, we introduce 2 additional metrics Information Density and NetScore from \cite{netscore} in \autoref{fig:radar} (Down). These two supplementary metrics help demonstrate a generalized reflection on performance and efficiency combined, making it easier to differentiate models. These data display a profound breakthrough of convolution-powered architecture and pure algorithm excellence in entire sign language translation field and Edge AI.

\subsection{Qualitative Comparison}
To ensure straightforward comparisons, we utilized GASLT’s qualitative table’s data points and performed direct comparisons in the following \autoref{tab: qual}. We listed Reference (GroundTruth), TSPNet \cite{tspnet}, GASLT, and results generated by our Signformer model. In the first example, our Signformer produces a slight mistake with the day prediction (3rd instead of 17th) while being accurate in most words and grammar. In contrast, TSPNet predicts both the date and day incorrectly, and GASLT is entirely correct. In the second example, our model accurately predicts the translation in whole, while TSPNet produces a severe translation error, and GASLT mistakenly escalates "warnungen" (warnings) to "unwetterwarnungen" (severe weather warnings), although contextually acceptable but problematic in other potential situations. In the last example, our model is again correct in all expression. GASLT has generated missing and inaccurate words, resulting in incomplete expression. TSPNet falsely interprets the day, temperature, and number of degrees.

\begin{table}[]
\centering
\caption{\textbf{Qualitative Results on translation, comparing against TSPNet and GASLT,} we allocated GASLT's  table's data points and perform translation tasking such that straightforward quality difference is guaranteed. \textcolor{red}{Red means semantically wrong, and (\_) stands for missing,} \textcolor{mygreen}{Blue means perfect translation}.}
\scriptsize
\setlength{\tabcolsep}{2pt} 

\renewcommand{\arraystretch}{0.9} 

\begin{tabularx}{\columnwidth}{@{}>{\hsize=0.15\hsize}X>{\hsize=0.8\hsize}X@{}}
\toprule
\textbf{Reference} & und nun die wettervorhersage für morgen donnerstag den siebzehnten dezember. \par
(and now the weather forecast for tomorrow, Thursday, December 17th.) \\ 
\textbf{TSPNet} & und \textcolor{red}{(\_)} die wettervorhersage für morgen donnerstag den \textcolor{red}{sechzehnten januar}.
\par
(and \textcolor{red}{\_}  the weather forecast for tomorrow, thursday the \textcolor{red}{sixteenth of january}.) \\
\textbf{GASLT} & \textcolor{mygreen}{und nun die wettervorhersage für morgen donnerstag den siebzehnten dezember.} \\
\textbf{Ours} & und nun die wettervorhersage für morgen donnerstag den \textcolor{red}{dritten} dezember \par(and now the weather forecast for tomorrow, Thursday, December \textcolor{red}{3rd}.) \\ \midrule
\textbf{Reference} & es gelten entsprechende warnungen des deutschen wetterdienstes. \par
(Corresponding warnings from the German weather service apply.) \\
\textbf{TSPNet} & \textcolor{red}{am montag gibt} es \textcolor{red}{hier und da schauer in der südwesthälfte viel sonne}. \par
(\textcolor{red}{On Monday there will be showers here and there and lots of sunshine in the southwest half.}) \\ 
\textbf{GASLT} & es gelten entsprechende. \textcolor{red}{ unwetterwarnungen} des deutschen wetterdienstes. \par (Corresponding 
\textcolor{red}{severe weather} warnings from the German weather service apply.) \\ 
\textbf{Ours} & \textcolor{mygreen}{es gelten entsprechende warnungen des deutschen wetterdienstes.} \\ \midrule
\textbf{Reference} & morgen reichen die temperaturen von einem grad im vogtland bis neun grad am oberrhein. (Tomorrow temperatures will range from one degree in the Vogtland to nine degrees on the Upper Rhine.) \\ 
\textbf{TSPNet} & \textcolor{red}{heute nacht zehn} grad \textcolor{red}{am oberrhein und fünf} grad am oberrhein. \par
(\textcolor{red}{Tonight ten} degrees \textcolor{red}{on the upper rhine} and \textcolor{red}{five} degrees on the upper rhine.) \\ 
\textbf{GASLT} & morgen \textcolor{red}{(\_) (\_)} temperaturen von \textcolor{red}{null (\_)} grad im vogtland bis neun grad am oberrhein. \par (Tomorrow temperatures \textcolor{red}{(\_) (\_)} from \textcolor{red}{zero} degrees in the Vogtland to nine degrees on the Upper Rhine.) \\ 
\textbf{Ours} & \textcolor{mygreen}{morgen reichen die temperaturen von einem grad im vogtland bis neun grad am oberrhein.} \\ \bottomrule
\end{tabularx}
\vspace*{-10pt}
\label{tab: qual}
\end{table}

%% file: 10_conclusion.tex
\section{Conclusion}
In this paper, we propose \textbf{Signformer}, a ground-up architectural innovation for deep learning in sign language, particularly for gloss-free translation. Signformer redefines the benchmark baseline for future research that leads to merge of both hard-hearing and common communities. We pushed extremities of model performance and efficiency to realize LLM-comparable competence but using exceptional succinct model architecture. Signformer surpasses Sign2GPT as the new 2nd place in gloss-free scoreboard, minimizing model complexity for nearly 1807 times fewer parameters (7B to 3.88M) than the finest SignLLM with approximate performance on BLEU and ROUGE. Our success stems from its pure end2end, from-scratch algorithmic design without dependence on any pretrained models, excess embedding source, advanced NLP tricks or additional datasets. Our lightest versions Signformer-Feather of 0.57M parameters and its CoPE-upgrade drastically outperform most SOTA approaches, mounting to Top 5 in the literature of 2024. We establish a scalable, Edge AI solution for practical, deployable, and sustainable use in real-world scenario, encouraging more efficient R\&D that accelerates the bridging movements between communities. We exploit convolution potential upon linguistic attributes with terminal feasibility for real-time responsive system in CPU-only environment or mobility. Signformer represents a profound advancement of the landscape, and we are looking forward to more capable extensions and a fully barrier-free world of communications.

%% file: 12_appendix.tex
\section{Appendix Section}
\label{sec:appendix_section}
Supplementary material goes here.